\documentclass[twocolumn]{bmcart}
\usepackage{amsfonts}
\usepackage{eurosym}
\usepackage{amsmath,graphicx}
\usepackage{amssymb}
\usepackage{lipsum}
\usepackage{algorithm}
\usepackage{algorithmic}
\usepackage{subcaption}
\usepackage{setspace}
\usepackage{epstopdf}
\usepackage{bigints}
\usepackage{setspace}
\usepackage{tikz}
\usepackage{xr}
\usepackage{xcolor,soul}
\usepackage[normalem]{ulem}
\usepackage{array}
\usepackage{amsthm,amsmath}
\usepackage[utf8]{inputenc}
\usepackage{multicol}
\usepackage{lipsum}

\startlocaldefs

\endlocaldefs
\begin{document}

\begin{frontmatter}

\begin{fmbox}
\dochead{Research}


\title{Optimal clustering with missing values}

\author[
   addressref={af1},                   
   noteref={n1},
   corref={af1},                        
   email={s.boluki@tamu.edu}   
]{\inits{SB}\fnm{Shahin} \snm{Boluki}}
\author[
   addressref={af1},
   noteref={n1},                       
   email={siamak@tamu.edu}
]{\inits{SZD}\fnm{Siamak} \snm{Zamani Dadaneh}}
\author[
   addressref={af1,af2},
   email={xqian@ece.tamu.edu}
]{\inits{XQ}\fnm{Xiaoning} \snm{Qian}}
\author[
   addressref={af1,af2},
   email={edward@ece.tamu.edu}
]{\inits{ERD}\fnm{Edward R.} \snm{Dougherty}}


\address[id=af1]{
  \orgname{Department of Electrical and Computer Engineering, Texas A\&M University}, 
  \street{MS3128 TAMU},                     %
  \postcode{77843},                                
  \city{College Station, TX},                              
  \cny{USA}                                    
}
\address[id=af2]{
  \orgname{TEES-AgriLife Center for Bioinformatics \& Genomic Systems Engineering}, 
  \postcode{77843},                                
  \city{College Station, TX},                              
  \cny{USA}                                    
}


\begin{artnotes}
\note[id=n1]{Equal contributor} 
\end{artnotes}



\begin{abstractbox}

\begin{abstract} 
\parttitle{Background} 
Missing values frequently arise in modern biomedical studies due to various reasons, including missing tests or complex profiling technologies for different omics measurements. Missing values can complicate the application of clustering algorithms, whose goals are to group points based on some similarity criterion. A common practice for dealing with missing values in the context of clustering is to first impute the missing values, and then apply the clustering algorithm on the completed data. 
\parttitle{Results} 
We consider missing values in the context of optimal clustering, which finds an optimal clustering operator with reference to an underlying random labeled point process (RLPP). We show how the missing-value problem fits neatly into the overall framework of optimal clustering by incorporating the missing value mechanism into the random labeled point process and then marginalizing out the missing-value process. In particular, we demonstrate the proposed framework for the Gaussian model with arbitrary covariance structures. Comprehensive experimental studies on both synthetic and real-world RNA-seq data show the superior performance of the proposed optimal clustering with missing values when compared to various clustering approaches.
\parttitle{Conclusion}
Optimal clustering with missing values obviates the need for imputation-based pre-processing of the data, while at the same time possessing smaller clustering errors. 
\end{abstract}


\begin{keyword}
\kwd{Clustering}
\kwd{Missing data}
\kwd{Optimal design}
\kwd{Pattern recognition}
\end{keyword}


\end{abstractbox}
\end{fmbox}

\end{frontmatter}

\section*{Background}

Clustering has been a mainstay of genomics since the early days of
gene-expression microarrays \cite{bendor}. For instance, expression
profiles can be taken over various tissue samples and then clustered
according to the expression levels for each sample, the aim being to
discriminate pathologies based on their differential patterns of gene
expression \cite{bittner}. In particular, model-based clustering,
which assumes that the data are generated by a finite mixture of underlying
probability distributions, has gained popularity over heuristic clustering
algorithms, for which there is no concrete way of determining the number of
clusters or the best clustering method \cite{yeung2001model}. Model-based clustering methods \cite{fraley2002model} provide more robust criteria for selecting the appropriate number of clusters. For example, in a Bayesian framework, utilizing Bayes Factor can incorporate both \emph{a priori} knowledge of different models, and goodness of fit of the parametric model to the observed data. Moreover, nonparametric models such as Dirichlet-process mixture models \cite{maceachern1998estimating} provide a more flexible approach for clustering, by automatically learning the number of components. 
In small-sample settings, model-based approaches that incorporate model uncertainty have proved successful in designing robust operators \cite{DaltonOBC,mahdin1,alireza2,mahdin2}, and in objective-based experiment design to expedite the discovery of such operators \cite{shahino1,ariana,shahino21}.

Whereas classification theory is grounded in feature-label distributions
with the error being the probability that the classifier mislabels a point \cite{DaltonOBC,alireza11};
clustering algorithms operate on random labeled point processes (RLPPs) with
error being the probability that a point will be placed into the wrong
cluster (partition) \cite{Dougherty2004}. An optimal (Bayes) clusterer
minimizes the clustering error and can be found with respect to an appropriate
representation of the cluster error \cite{dalton2015analytic}.

A common problem in clustering is the existence of missing values. These are
ubiquitous with high-throughput sequencing technologies, such as
microarrays \cite{schena1995quantitative} and RNA sequencing (RNA-seq) \cite%
{mortazavi2008mapping}. For instance, with microarrays, missing data can
occur due to poor resolution, image corruption, or dust or scratches on the
slide \cite{troyanskaya2001missing}, while for RNA-seq, the sequencing
machine may fail to detect genes with low expression levels owing to the
random sampling nature of sequencing technologies. As a result of these
missing data mechanisms, gene expression data from microarray or RNA-seq
experiments are usually in the form of large matrices, with rows and columns
corresponding to genes and experimental conditions or different subjects,
respectively, with some values missing. Imputation methods, such as \emph{%
MICE} \cite{buuren2010mice}, \emph{Amelia II} \cite{honaker2011amelia} and 
\emph{missForest} \cite{stekhoven2011missforest}, are usually employed to
complete the data matrix before clustering analysis; however, in
small-sample settings, which are common in genomic applications, these
methods face difficulties, including co-linearity due to potential high correlation
between genes in samples, which precludes the successful imputation of missing
values.

In this paper we follow a different direction by incorporating the
generation of missing values with the original generating random labeled
point process, thereby producing a new RLPP that generates the actual
observed points with missing values. The optimal clusterer in the context of missing values is
obtained by marginalizing out the missing features in the new RLPP. One
potential challenge arising here is that in the case of missing values with
general patterns, conducting the marginalization can be computationally
intractable, and hence resorting to approximation methods such as Monte
Carlo integration is necessary.

Although the proposed framework for optimal clustering can incorporate the
probabilistic modeling of arbitrary types of missing data mechanisms, to
facilitate analysis, throughout this work we assume data are missing
completely at random (MCAR) \cite{little2014statistical}. In this scenario,
the parameters of the missingness mechanism are independent of other model
parameters and therefore vanish after the expectation operation in the
calculation of the posterior of label functions for clustering assignment.

We derive the optimal clusterer for different scenarios in which features
are distributed according to multivariate Gaussian distributions. The
performance of this clusterer is compared to various methods, including $k$-POD \cite{kpod} and fuzzy $c$-means with optimal completion strategy \cite{fcm-ocs}, which are methods for directly clustering data with missing values, and also $k$%
-means \cite{kanungo2002efficient}, fuzzy $c$-means \cite{bezdek1984fcm} and
hierarchical clustering \cite{johnson1967hierarchical} with the missing values imputed. Comprehensive
simulations based on synthetic data show the superior performance of the
proposed framework for clustering with missing values over a range of
simulation setups. Moreover, evaluations based on RNA-seq data further
verify the superior performance of the proposed method in a real-world
application with missing data.

\section*{Methods}

\subsection*{Optimal clustering}

Given a point set $S \subset \mathbb{R}^d$, where $d$ is the dimension of
the space, denote the number of points in $S$ by $\eta(S)$. A \emph{random
labeled point process} (RLPP) is a pair $(\Xi,\Lambda)$, where $\Xi$ is a
point process generating $S$ and $\Lambda$ generates random labels on point
set $S$. $\Xi$ maps from a probability space to $[N;\mathcal{N}]$, where $N$
is the family of finite sequences in $\mathbb{R}^d$ and $\mathcal{N}$ is the
smallest $\sigma$-algebra on $N$ such that for any Borel set $B$ in $\mathbb{%
R}^d$, the mapping $S \rightarrow \eta(S \cap B)$ is measurable. A random
labeling is a family, $\Lambda = \{ \Phi_S: S \in N \}$, where $\Phi_S$ is a
random label function on the point set $S$ in $N$. Denoting the set of
labels by $L=\{1,2,...,l\}$, $\Phi_S$ has a probability mass function on $%
L^S $ defined by $P_S(\phi_S)=P(\Phi_S=\phi_S|\Xi=S)$, where $\phi_S:S
\rightarrow L$ is a deterministic function assigning a label to each point
in $S$.

A label operator $\lambda$ maps point sets to label functions, $%
\lambda(S)=\phi_{S,\lambda} \in L^S$. For any set $S$, label function $%
\phi_S $ and label operator $\lambda$, the \emph{label mismatch error} is
defined as

\begin{equation}
\epsilon _{\lambda }(S,\phi _{S})=\frac{1}{\eta (S)}\sum_{x\in S}I_{\phi
_{S}(x)\neq \phi _{S,\lambda }(x)},
\end{equation}%
where $I_{A}$ is an indicator function equal to 1 if $A$ is true and 0
otherwise. The \emph{error of label function} $\lambda (S)$ is computed as $%
\epsilon _{\lambda }(S)=\mathbb{E}_{\Phi _{S}}[\epsilon _{\lambda }(S,\phi
_{S})|S]$, and the \emph{error of label operator} $\lambda$ for the corresponding RLPP is then defined
by $\epsilon \lbrack \lambda ]=\mathbb{E}_{\Xi }\mathbb{E}_{\Phi _{\Xi
}}[\epsilon _{\lambda }(\Xi ,\phi _{\Xi })]$.

Clustering involves identifying partitions of a point set rather than the
actual labeling, where a partition of $S$ into $l$ clusters has the form $%
\mathcal{P}_S = \{S_1,S_2,...,S_l \}$ such that $S_i$'s are disjoint and $%
S = \bigcup_{i=1}^l S_i$. A cluster operator $\zeta$ maps point sets to
partitions, $\zeta(S)=\mathcal{P}_{S,\zeta}$. Considering the label
switching property of clustering operators, let us define $F_{\zeta}$ as the
family of label operators that all induce the same partitions as the
clustering operator $\zeta$. More precisely, a label function $\phi_S$
induces partition $\mathcal{P}_S = \{S_1,S_2,...,S_l \}$, if $S_i = \{ x \in
S : \phi_S(x)=l_i \}$ for distinct $l_i \in L$. Thereby, $\lambda \in
F_{\zeta}$ if and only if $\phi_{S,\lambda}$ induces the same partition as $%
\zeta(S)$ for all $S \in N$. For any set $S$, label function $\phi_S$ and
cluster operator $\zeta$, define the \emph{cluster mismatch error} by

\begin{equation}
\epsilon_{\zeta}(S,\phi_S) = \min_{\lambda \in F_{\zeta}}
\epsilon_{\lambda}(S,\phi_S) ,
\end{equation}
the \emph{error of partition} $\zeta(S)$ by $\epsilon_{\zeta}(S) = \mathbb{E}%
_{\Phi_S} [\epsilon_{\zeta}(S,\phi_S)|S ]$ and the \emph{error of cluster
operator} $\zeta$ for the RLPP by $\epsilon[\zeta] = \mathbb{E}_{\Xi} \mathbb{E}%
_{\Phi_{\Xi}} [\epsilon_{\zeta}(\Xi,\phi_{\Xi})]$.


As shown in \cite{dalton2015analytic}, error definitions for partitions can
be represented in terms of risk with intuitive cost functions. Specifically,
define $G_{\mathcal{P}_S}$ such that $\phi_S \in G_{\mathcal{P}_S}$ if and
only if $\phi_S$ induces $\mathcal{P}_S$. The error of partition can be
expressed as

\begin{equation}
\epsilon_{\zeta}(S) = \sum_{\mathcal{P}_S \in \mathcal{K}_S} c_S(\zeta(S),%
\mathcal{P}_S) P_S(\mathcal{P}_S),
\end{equation}
where $\mathcal{K}_S$ is the set of all possible partitions of $S$, $P_S(%
\mathcal{P}_S) = \sum_{\phi_S \in G_{\mathcal{P}_S}} P_S(\phi_S)$ is the
probability mass function on partitions $\mathcal{P}_S$ of $S$, and the 
\emph{partition cost function} between partitions $\mathcal{P}_S$ and $\mathcal{Q}_S$ of $S$ is defined as

\begin{equation}
c_{S}(\mathcal{Q}_{S},\mathcal{P}_{S})=\frac{1}{\eta (S)}\min_{\phi _{S,%
\mathcal{Q}_{S}}\in G_{\mathcal{Q}_{S}}}\sum_{x\in S}I_{\phi _{S,\mathcal{P}%
_{S}}\neq \phi _{S,\mathcal{Q}_{S}}},
\end{equation}%
with $\phi _{S,\mathcal{P}_{S}}$ being any member of $G_{\mathcal{P}_{S}}$.
A Bayes cluster operator $\zeta ^{\ast }$ is a clusterer with the minimal error $%
\epsilon \lbrack \zeta ^{\ast }]$, called the \emph{Bayes error}, obtained
by a Bayes partition, $\zeta ^{\ast }(S)$ for each set $S\in N$:

\begin{eqnarray}
\zeta^*(S) &=& \arg \min_{\zeta(S) \in \mathcal{K}_S} \epsilon_{\zeta}(S) 
\notag \\
&=& \arg \min_{\zeta(S) \in \mathcal{K}_S} \sum_{\mathcal{P}_S \in \mathcal{K%
}_S} c_S(\zeta(S),\mathcal{P}_S) P_S(\mathcal{P}_S).  \notag \\
\end{eqnarray}
The Bayes clusterer can be solved for each fixed $S$ individually. More specifically, the search space in the minimization and the set of partitions with known probabilities in the summation can be constrained to subsets of $\mathcal{K}_S$, denoted by $\mathcal{C}_S$ and $\mathcal{R}_S$, receptively. We refer to $\mathcal{C}_S$ and $\mathcal{R}_S$ as the set of candidate partitions and the set of reference partitions, respectively. Following \cite{dalton2015analytic}, we can search for the optimal clusterer based on both optimal and suboptimal procedures (detailed in Results and discussion section) with derived bounds that can be used to optimally reduce the size of $\mathcal{C}_S$ and $\mathcal{R}_S$.

\subsection*{Gaussian model with missing values}

We consider an RLPP model that generates the points in the set $S$ according to
a Gaussian model, where features of $x \in S$ can be missing completely at
random due to a missing data mechanism independent of the RLPP. More
precisely, the points $x \in S$ with label $\phi_S(x) = i$ are drawn
independently from a Gaussian distribution with parameters $\rho_i =
\{\mu_i,\Sigma_i\} $. Assuming $n_i$ sample points with label $i$, we divide
the observations into $G_i \leq n_i$ groups, where all $n_{ig}$ points in
group $g$ have the same set, $J_{ig}$, of observed features with cardinality 
$|J_{ig}| = d_{ig}$. Denoting by $S_{ig}$ the set of sample points in group $%
g$ of label $i$, we represent the pattern of missing data in this group
using a $d_{ig}\times d$ matrix $M_{ig}$, where each row is a $d$%
-dimensional vector with a single non-zero element with value 1
corresponding to the observed feature's index. Thus, the non-missing portion
of sample point $x \in S_{ig}$, i.e. $M_{ig}x$, has Gaussian distribution $%
\text{N}(M_{ig}\mu_i ,M_{ig}\Sigma_i M_{ig}^{T})$.

Given $\rho =\{\rho _{1},\rho _{2},...,\rho _{l}\}$ of
independent parameters, to evaluate the posterior probability of random
labeling function $\phi _{S}\in L^{S}$, we have 
\begin{align}
P_{S}& (\phi _{S})\propto P(\phi _{S})f(S|\phi _{S})=  \notag  \label{eq:ps}
\\
& P(\phi _{S})\int f(S|\phi _{S},\rho)f(\rho )d\rho =  \notag \\
& P(\phi _{S})\prod_{\substack{ i=1  \\ n_{i}\geq 1}}^{l}\int \Big(%
\prod_{x\in S_{i}}f_{i}(x|\rho _{i})\Big)f(\rho _{i})d\rho _{i}=  \notag \\
& P(\phi _{S})\prod_{\substack{ i=1  \\ n_{i}\geq 1}}^{l}\int \Big(%
\prod_{g=1}^{G_{i}}\prod_{x\in S_{ig}} \\
& \mbox{N}\big(M_{ig}x;M_{ig}\mu _{i},M_{ig}\Sigma _{i}M_{ig}^{T}\big)\Big)%
f(\mu _{i},\Sigma _{i})d\mu _{i}d\Sigma _{i},  \notag
\end{align}%
%
where $P(\phi _{S})$ is the prior probability on label functions, which we
assume does not depend on the specific points in $S$. 

\subsubsection*{Known means and covariances}

When mean and covariance parameters of label-conditional distributions are
known, the prior probability $f(\mu_i,\Sigma_i)$ in~(\ref{eq:ps}) is a point
mass at $\rho_i = \{\mu_i,\Sigma_i\} $. Thus,
\begin{equation}  \label{eq:ps1}
\begin{split}
&P_S(\phi_S)\propto P(\phi_S) \times   \\
&\prod_{\substack{ i=1  \\ n_i \geq 1}}^l \prod_{g=1}^{G_i} \prod_{x \in
S_{ig}} \Big[ (2\pi)^{-d_{ig}/2} |M_{ig}\Sigma_i M_{ig}^{T}|^{-1/2} \times \\
&\exp \big\{ - \frac{1}{2} (x-\mu_i)^T M_{ig}^{T} (M_{ig}\Sigma_i
M_{ig}^{T})^{-1} M_{ig} (x-\mu_i) \big\} \Big].  
\end{split}
\end{equation} 
We define the group-$g$ statistics of label $i$ as 
\begin{eqnarray}
m_{ig} &:=&\frac{1}{n_{ig}}\sum_{x \in S_{ig}}M_{ig}x,  \notag \\
\Psi_{ig} &:=&\sum_{x \in S_{ig}}(M_{ig}x-m_{ig})(M_{ig}x-m_{ig})^{T},
\label{eq:def}
\end{eqnarray}
where $m_{ig}$ and $\Psi_{ig}$ are the sample mean and scatter matrix,
employing only the observed $n_{ig}$ data points in group $g$ of label $i$.
The posterior probability of labeling function~(\ref{eq:ps1}) then can be
expressed as

\begin{equation}  \label{eq:ps2}
\begin{split}
&P_S(\phi_S)\propto P(\phi_S) \prod_{\substack{ i=1  \\ n_i \geq 1}}^l
\prod_{g=1}^{G_i} \\
&\Big[|2\pi \Sigma_{ig}|^{-n_{ig}/2}\exp \{- \frac{1}{2}\text{tr}\big(%
\Psi_{ig}(\Sigma _{ig})^{-1}\big)\} \times \\
&\exp \big\{-\frac{1}{2}n_{ig}(m_{ig}-M_{ig}\mu_i)^{T}(%
\Sigma_{ig})^{-1}(m_{ig}-M_{ig}\mu_i )\big\} \Big],
\end{split}
\end{equation}
where $\Sigma_{ig}=M_{ig}\Sigma_i M_{ig}^{T}$ is the covariance matrix
corresponding to group $g$ of label $i$.

\subsubsection*{Gaussian means and known covariances}

Under this model, data points are generated according to Gaussians whose
mean parameters are random and their covariance matrices are fixed.
Specifically, for label $i$ we have $\mu_i \sim \mbox{N}(m_i,\frac{1}{\nu_i}%
\Sigma_i)$, where $\nu_i>0$ and $m_i$ is a length $d$ real vector. Thus the
posterior of label function given the point set $S$ can be derived as
\begin{equation}  \label{eq:ps3}
\begin{split}
&P_S(\phi_S) \propto P(\phi_S) \prod_{\substack{ i=1  \\ n_i \geq 1}}^l %
\Bigg[ \prod_{g=1}^{G_i} \Big[ |2\pi\Sigma_{ig}|^{-n_{ig}/2} \times 
\\
&\exp \{- \frac{1}{2}\text{tr}\big(\Psi_{ig}(\Sigma _{ig})^{-1}\big)\} %
\Big] \times (\nu_i)^{d/2} |2\pi \Sigma_i|^{-1/2}  \\
&\int \exp \Big\{-\frac{1}{2}\sum_{g=1}^{G_i}n_{ig}(m_{ig}-M_{ig}%
\mu_i)^{T}(\Sigma_{ig})^{-1} \\
&(m_{ig}-M_{ig}\mu_i ) - \frac{\nu_i}{2} (\mu_i-m_i)^T \Sigma_i^{-1}
(\mu_i-m_i) \Big\} d\mu_i \Bigg]. 
\end{split}
\end{equation}

By completing the square and using the normalization constant of
multivariate Gaussian distribution, the integral in this equation can be
expressed as
\begin{align}
& \int \exp \Big\{-\frac{1}{2}\big[(\mu _{i}-A_{i}^{-1}b_{i})^{T}A_{i}(\mu
_{i}-A_{i}^{-1}b_{i})+  \notag \\
& \sum_{g=1}^{G_{i}}n_{ig}m_{ig}^{T}\Sigma _{ig}^{-1}m_{ig}+\nu
_{i}m_{i}^{T}\Sigma _{i}^{-1}m_{i}-b_{i}^{T}A_{i}^{-1}b_{i}\big]\Big\} 
\notag \\
& =|A_{i}/(2\pi )|^{-1/2}\exp \Big\{-\frac{1}{2}\big[%
\sum_{g=1}^{G_{i}}n_{ig}m_{ig}^{T}\Sigma _{ig}^{-1}m_{ig}+ \\
& \quad \quad \quad \quad \quad \quad \nu _{i}m_{i}^{T}\Sigma
_{i}^{-1}m_{i}-b_{i}^{T}A_{i}^{-1}b_{i}\big]\Big\},  \notag
\end{align}%
where%
\protect\begin{eqnarray}
A_{i} &=&\sum_{g=1}^{G_{i}}n_{ig}M_{ig}^{T}\Sigma _{ig}^{-1}M_{ig}+\protect%
\nu _{i}\Sigma _{i}^{-1}, \\
b_{i} &=&\sum_{g=1}^{G_{i}}n_{ig}M_{ig}^{T}\Sigma _{ig}^{-1}m_{ig}+\protect%
\nu _{i}\Sigma _{i}^{-1}m_{i}. 
\protect\end{eqnarray}%

\subsubsection*{Gaussian-Inverse-Wishart Means and Covariances}

Under this model, data points are generated from Gaussian distributions with
random mean and covariance parameters. More precisely, the parameters
associated with label $i$ are distributed as ${\mu_i|\Sigma_i \sim \mbox{N}(m_i,\frac{1%
}{\nu_i}\Sigma_i)}$ and $\Sigma_i \sim \mbox{IW}(\kappa_i,\Psi_i)$, where the
covariance has inverse-Wishart distribution

\begin{equation}
f(\Sigma_i) = \frac{|\Psi_i|^{\kappa_i/2}}{2^{\kappa_i d/2}
\Gamma_d(\kappa_i/2)} |\Sigma_i|^{\frac{\kappa_i+d+1}{2}} \exp \big( -\frac{1%
}{2} \mbox{tr}(\Psi_i \Sigma_i^{-1}) \big).
\end{equation}

To compute the posterior probability of labeling function (\ref{eq:ps}), we
first marginalize out the mean parameters $\mu_i$ in a similar fashion to (%
\ref{eq:ps3}), obtaining

\begin{eqnarray}  \label{eq:iw}
P_S(\phi_S) &\propto& P(\phi_S) \prod_{\substack{ i=1  \\ n_i \geq 1}}^l
\int \Bigg[ \prod_{g=1}^{G_i}|2\pi \Sigma_{ig}|^{-n_{ig}/2}\times  \notag \\
&&\exp \{- \frac{1}{2}\text{tr}\big(\Psi_{ig}(\Sigma _{ig})^{-1}\big)\}\times
\\
&& (\nu_i)^{d/2} |\Sigma_i|^{-1/2} |A_i/(2\pi)|^{-1/2}\times  \notag \\
&& \exp \Big\{ -\frac{1}{2} \big[ \sum_{g=1}^{G_i}n_{ig} m_{ig}^T
\Sigma_{ig}^{-1} m_{ig}  \notag \\
&+& \nu_i m_i^T \Sigma_i^{-1} m_i - b_i^T A_i^{-1} b_i \big] \Big\} \Bigg] %
f(\Sigma_i) d\Sigma_i.  \notag
\end{eqnarray}

The integration in the above equation has no closed form solution, thus we
resort to Monte Carlo integration for approximating it. Specifically,
denoting the term in the brackets in equation~(\ref{eq:iw}) as $g(\Sigma_i)$%
, we draw $J$ samples $\Sigma_i^{(j)} \sim \mbox{IW}(\kappa_i,\Psi_i)$, $%
j=1,2,...,J$, and then compute the integral as $\frac{1}{J} \sum_{j=1}^{J}
g(\Sigma_i^{(j)})$.

\section*{Results and discussion}

The performance of the proposed method for optimal clustering with missing
values at random is compared with some suboptimal versions, two other methods for clustering data with missing values, and also classical clustering algorithms with imputed missing values. The performance comparison is carried out on
synthetic data generated from different Gaussian RLPP models with different
missing probability setups, and also on a publicly available dataset of
breast cancer generated by TCGA Research Network (https://cancergenome.nih.gov/). In our experiments, the
results of the exact optimal solution for the RLPP with missing at random
(Optimal) is provided for smaller point sets, i.e. wherever computationally
feasible. We have also tested two suboptimal solutions, similar to the ideas
in \cite{dalton2015analytic}, for an RLPP with missing at random. In the
first one (Subopt. Pmax), the set of reference partitions ($\mathcal{R}_S$) is restricted to a
closed ball of a specified radius centered on the
partition with the highest probability, where the distance of two partitions is defined as the minimum Hamming distance between labels inducing the partitions. In both Optimal and Pmax, the
reference set is further constrained to the partitions that assign the correct
number of points to each cluster, but the set of candidate partitions ($\mathcal{C}_S$) includes all the possible partitions of $n$ points, i.e. $2^{n-1}$. In the second suboptimal solution (Subopt. Pseed), a local search within Hamming
distance at 1 is performed starting from five random initial partitions to
approximately find the partition with (possibly local) maximum probability.
Then the sets of reference and candidate partitions are constrained to the
partitions with correct cluster sizes with a specified Hamming distance from
the found (local) maximum probability partition. The bounds derived in \cite%
{dalton2015analytic} for reducing the set of candidate and reference
partitions are used, where applicable, in Optimal, Pseed, and Pmax.

In all scenarios, $k$-POD and fuzzy $c$-means with optimal completion strategy (FCM-OCS) are directly applied to the data with missing values. In the simulations in \cite{fcm-ocs}, where FCM-OCS is introduced, to initialize cluster centers, the authors apply ordinary fuzzy $c$-means to the complete data, i.e. using knowledge of the missing values. To have a fair comparison with other methods, we calculate the initial cluster centers for FCM-OCS by applying fuzzy $c$-means to the subset of points with no missing features for lower missing rates. For higher missing rates we impute the missing values by the mean of the corresponding feature values across all points, and then apply fuzzy $c$-means to all the points to initialize the cluster centers. In order to apply the classical algorithms, the missing
values are imputed according to \cite{siamak}, by employing a multivariate
Gibbs sampler that iteratively generates samples for missing values and
parameters based on the observed data. The classical algorithms included in
our experiments include \emph{k}-means (KM), fuzzy \emph{c}-means (FCM),
hierarchical clustering with single linkage (Hier. (Si)), and hierarchical
clustering with complete linkage (Hier. (Co)). Moreover, completely random
clusterer (Random) results are also included for performance comparisons.

\subsection*{Simulated data}

In the simulation analysis, the number of clusters is fixed at 2, and the
dimensionality of the RLPPs (number of features) is set to 5. Additional results for 20 features are provided in Additional file 1. Point
generation is done based on a Gaussian mixture model (GMM). Three different
scenarios for the parameters of the GMM are considered: \emph{i}) Fixed
known means and covariances \emph{ii}) Known covariances and unknown means
with Gaussian distributions. \emph{iii}) Unknown means and covariances with
Gaussian inverse-Wishart distributions. We select the values of the
parameters of the point generation process to have an approximate Bayes
error of 0.15. The selected values are shown in Table \ref{table:sim-params}%
. 
\begin{table*}[tph]
\caption{Parameters for the point generation under three models. $\mbox{N}$, 
$\mbox{IW}$, $\mathbf{1}_{d}$ , and $I_{d}$ denote Gaussian,
inverse-Wishart, column vector of all ones with length $d$, and $d\times d$
idendity matrix, respectively.}
\label{table:sim-params}
\begin{center}
\resizebox{0.99\linewidth}{!} {\begin{tabular}{|l|l|l|l|}
\hline
Model & Mean vectors & Covariance matrices & Distributions' hyperparameters
\\ \hline
&  &  &  \\ 
Fixed means and covariances & $\mu_1=0\cdot \mathbf{1}_d$, $\mu_2=0.445\cdot 
\mathbf{1}_d$ & $\Sigma_1=\Sigma_2=0.23\cdot I_d$ & --- \\ \hline
Gaussian means and fixed covariances & $\mu_1 \sim \mbox{N}(m_1,\frac{1}{\nu_1}\Sigma_1)$, $\mu_2 \sim \mbox{N}(m_2,\frac{1}{\nu_2}\Sigma_2)$ & $\Sigma_1=\Sigma_2=0.28\cdot I_d$ & $m_1=0\cdot \mathbf{1}_d$, $m_2=0.45\cdot 
\mathbf{1}_d$, \\ 
&  &  & $\nu_1=30$, $\nu_2=5$ \\ \hline
Gaussian means and inverse-Wishart covariances & $\mu_1 \sim \mbox{N}(m_1,\frac{1}{\nu_1}\Sigma_1)$, $\mu_2 \sim \mbox{N}(m_2,\frac{1}{\nu_2}\Sigma_2)$
& $\Sigma_1\sim \mbox{IW}(\kappa_1,\Psi_1)$,$\Sigma_2\sim \mbox{IW}(\kappa_2,\Psi_2)$ & $m_1=0\cdot \mathbf{1}_d$, $m_2=0.45\cdot \mathbf{1}_d$,
\\ 
&  &  & $\nu_1=30$, $\nu_2=5$, \\ 
&  &  & $\Psi_1=\Psi_2=20.7\cdot I_d$, \\ 
&  &  & $\kappa_1=\kappa_2=75$ \\ \hline
\end{tabular}} 
\end{center}
\end{table*}
For the point set generation, the number of points from each cluster is
fixed \emph{a priori}. The distributions are first drawn from the assumed
model, and then the points are generated based on the drawn distributions. A
subset of the points' features is randomly selected to be hidden based on
missing at random with different missing probabilities. Four different
setups for the number of points are considered in our simulation analysis:
10 points from each cluster ($n_{1}=n_{2}=10$), 12 points from one cluster
and 8 points from the other cluster ($n_{1}=12,n_{2}=8$), 35 points from
each cluster ($n_{1}=n_{2}=35$), and 42 points from one cluster and 28
points from the other cluster ($n_{1}=42,n_{2}=28$). When having unequal
sized clusters, in half of the repetitions $n_{1}$ points are generated from
the first distribution and $n_{2}$ points from the second distribution, and
vice-versa in the other half. In each simulation repetition, all clustering
methods are applied to the points to generate a vector of labels that
induces a two-cluster partition. The predicted label vector by each method
is compared with the true label vector of each point in the point set to
calculate the error of that method on that point set. In other words, for
each method the number of points assigned to a cluster different from their
true one are counted (after accounting for the label switching issue) and
divided by the total number of points ($n=n_{1}+n_{2}$) to calculate the
clustering error of that method on the point set. These errors are averaged
across all point sets in different repetitions to empirically estimate the
clustering error of each method under a model and fixed missing-value
probability. In cases with $n=70$, since applying Optimal and Pmax is
computationally prohibitive, we only provide the results for Pseed.

In Additional file 1, the average clustering errors are shown as a function
of the Hamming distance threshold used to define the set of reference
partitions in Pmax and Pseed, for different simulation scenarios. From the
Figures in Additional file 1, we see that in all cases, the
performances of Pmax and Pseed are quite insensitive to the set threshold of
the Hamming distance for reference partitions. Note that in these types of
figures all the other methods' performances other than Pmax and Pseed are
constant in each plot.

\begin{figure*}[tph]
\centering
\includegraphics[width=0.99\textwidth]{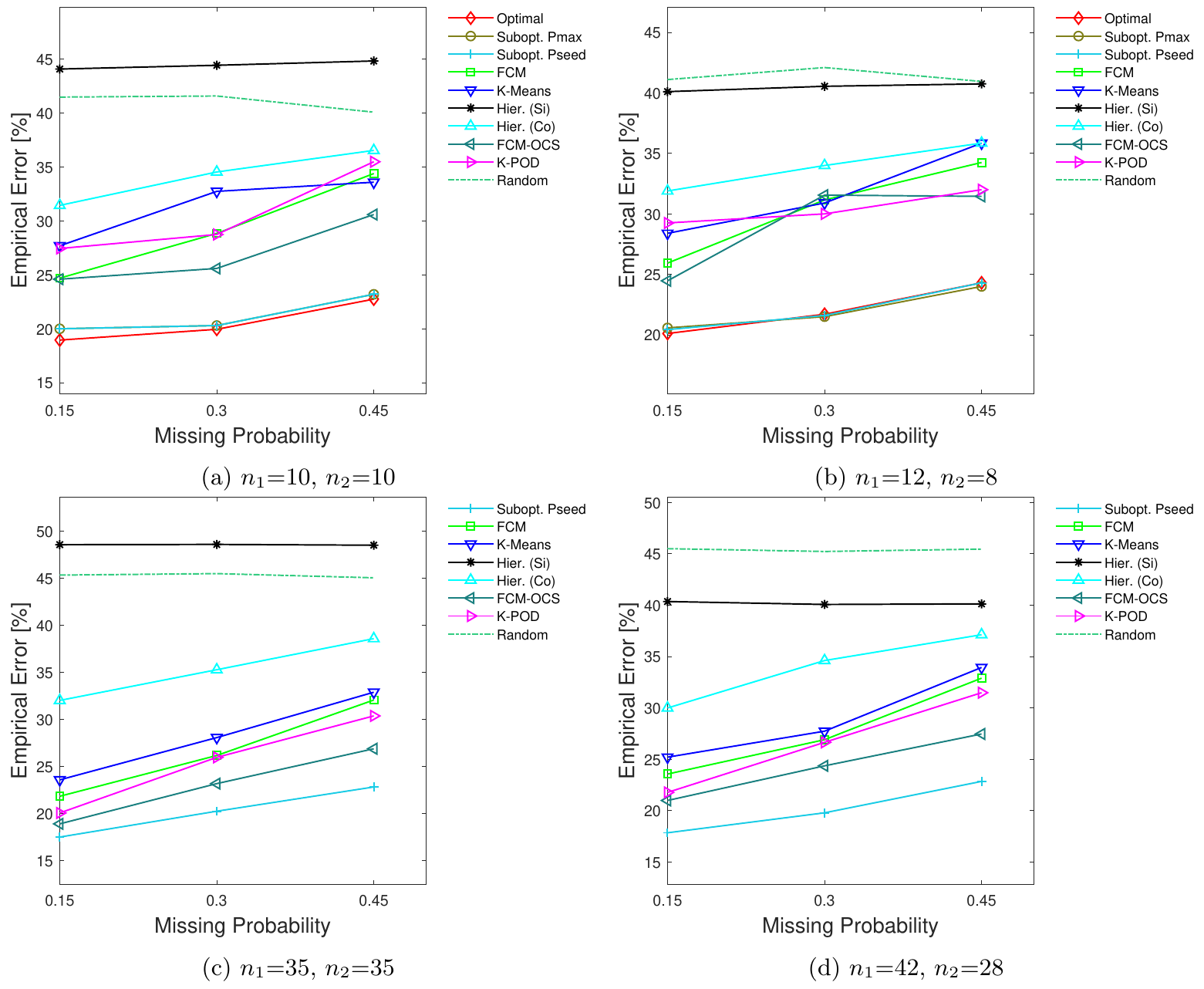}
\caption{\csentence{Average clustering
errors
vs.
missing
probability
for
fixed means and covariances model.}
The first and second rows correspond to $n=20$ and $n=70$, respectively.}
\label{fig:sim-fixedmeanfixedcov}
\end{figure*}

The average results for the fixed mean vectors and covariance matrices
across 100 repetitions are shown in Figure \ref{fig:sim-fixedmeanfixedcov}.
Here, the Hamming distance threshold for reference partitions in Pmax and
Pseed is fixed at 1. It can be seen that Optimal, Pmax, and Pseed outperform
all the other methods in all the smaller sample size settings, and Pmax
and Pseed have virtually the same performance as Optimal. For the larger
sample size settings where only Pseed is applied, its superior performance
compared with other methods is clear from the figure.

\begin{figure*}[tph]
\centering
\includegraphics[width=0.99\textwidth]{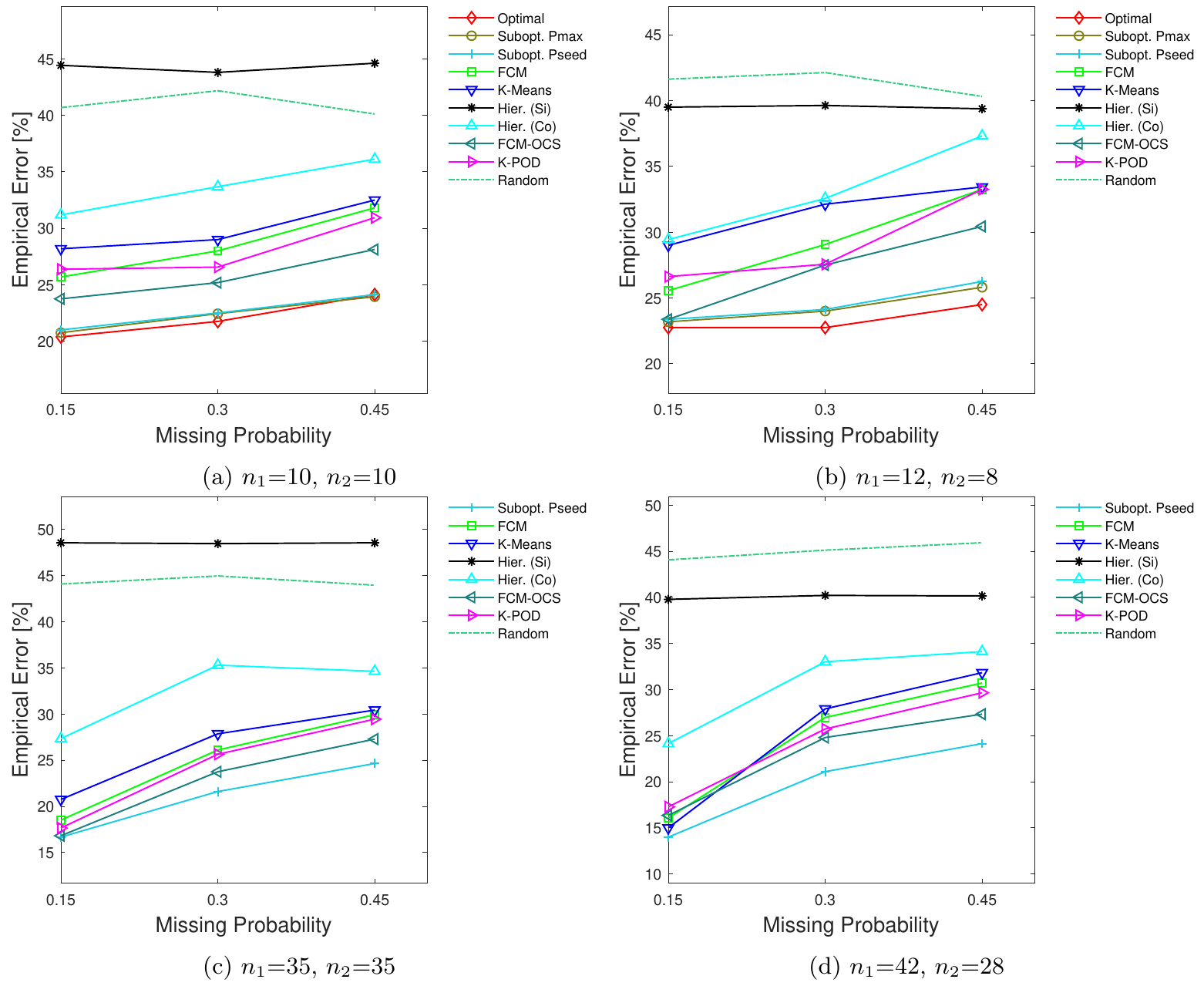}
\caption{\csentence{Average clustering
errors
vs.
missing
probability
for
Gaussian means and fixed covariances
model.} The first and second rows correspond to $n=20$ and $n=70$,
respectively.}
\label{fig:sim-gaussianmeanfixedcov}
\end{figure*}

Figure \ref{fig:sim-gaussianmeanfixedcov} depicts the comparison results
under the unknown mean vectors with Gaussian distributions and fixed
covariance matrices averaged over 80 repetitions. The Hamming distance
threshold in Pmax and Pseed is set to 2. For smaller sample sizes, Optimal,
Pmax and Pseed have lower average errors than all the other methods. We can
see that for balanced clusters the suboptimal and optimal solutions have
very close performances, but for the unbalanced clusters case with higher
missing probabilities the difference between Optimal and Pmax and Pseed gets
noticeable. For larger sample sizes Pseed consistently outperforms the other
methods, although for lower missing probabilities it has closer competitors.
In all cases, as the missing probability increases, the superior performance
of the proposed methods becomes more significant. 

\begin{figure*}[tph]
\centering
\includegraphics[width=0.99\textwidth]{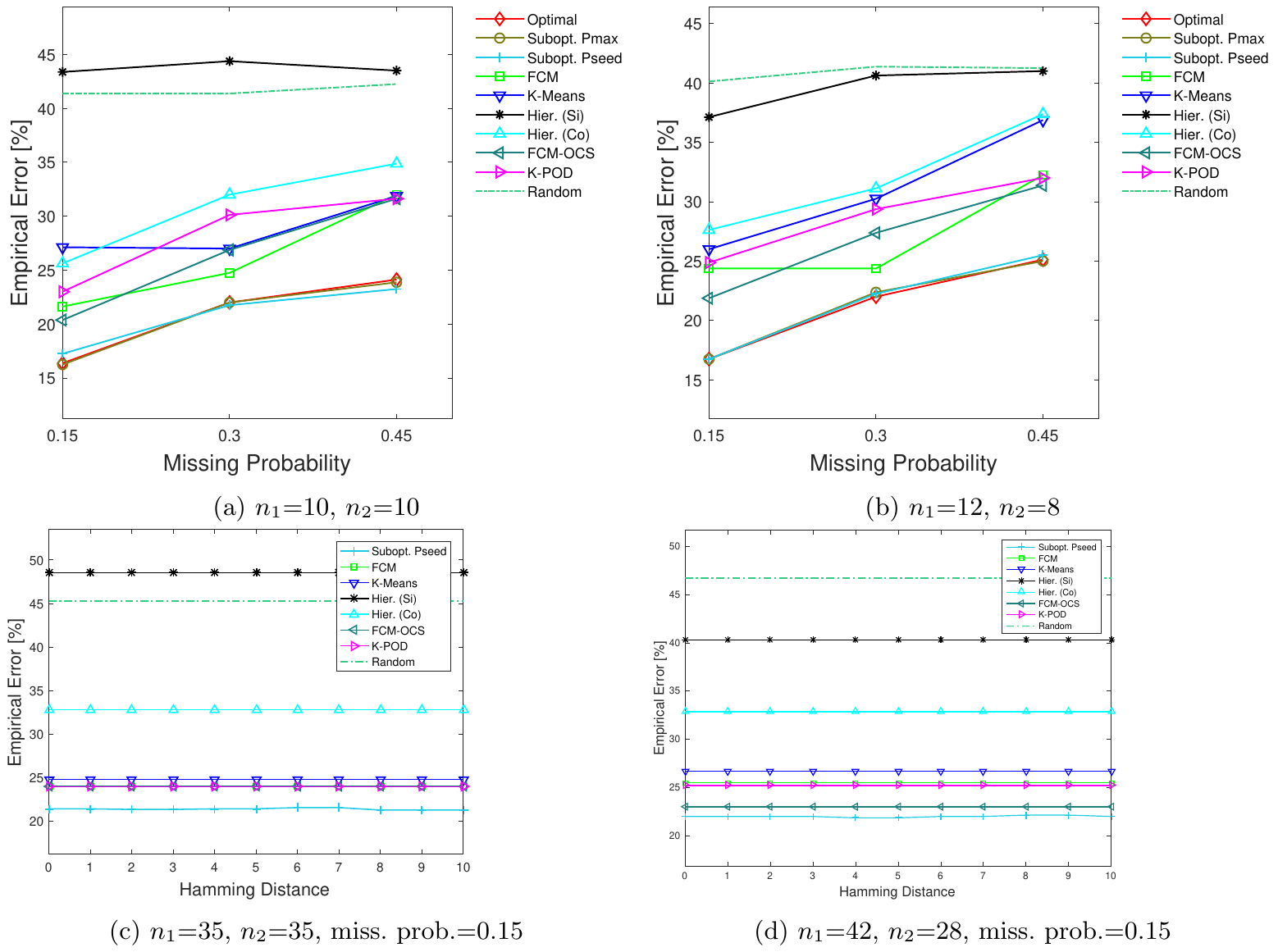}
\caption{\csentence{Average clustering
errors
for
Gaussian
means
and
inverse-Wishart covariances model.} The
first row corresponds to $n=20$, and the errors are shown for different
missing probabilities. The second row corresponds to $n=70$ and missing
probability of 0.15, where the errors are plotted vs. the Hamming distance
threshold used to define the reference partitions in Pseed.}
\label{fig:sim-gaussianmeaninversewishartcov}
\end{figure*}

The average results under the unknown mean vectors and coavriance matrices
with Gaussian-inverse-Wishart distribution over 40 repetitions are provided
in Figure \ref{fig:sim-gaussianmeaninversewishartcov}. In the smaller sample
size cases, the Hamming distance threshold in Pmax and Pseed is fixed at 8,
and we can see that the proposed suboptimal (Pmax and Pseed) and optimal
clustering with missing values have very close average errors, and
all are much lower than the other methods' errors. For larger sample sizes,
only the results for missing probability equal to 0.15 are shown vs. the
Hamming distance threshold used to define the reference partitions in Pseed. Again, Pseed performs better than the other methods. 

\subsection*{RNA-seq data}

In this section the performance of the clustering methods are examined on a
publicly available RNA-seq dataset of breast cancer. The data is available
on The Cancer Genome Atlas (TCGA) \cite{cancer2008comprehensive}, and is
procured by the R package TCGS2STAT \cite{wan2015tcga2stat}. It consists of
matched tumor and normal samples, and includes 97 points from each. The
original data are in terms of the number of sequence reads mapped to each
gene. RNA-seq data are integers, highly skewed and over-dispersed \cite%
{ehsan1,ehsan2,arianan1}. Thus, we apply a variance stabilizing
transformation (VST) \cite{durbin2002variance} implemented in DESeq2 package 
\cite{love2014moderated}, and transform data to a log2 scale that have been
normalized with respect to library size. For all subsequent analysis, other
than for calculating clustering errors, we assume that the labels of data
are unknown. Feature selection is performed in a completely unsupervised
manner, since in clustering no labels are known in practice. The top 10
genes in terms of variance to mean ratio of expression are picked as
features to be used in clustering algorithms. In general, for setting prior hyperparameters, external sources of information like signaling pathways, where available, can be leveraged \cite{shahinp1,shahinp2}. Here, we only use a subset of the discarded gene expressions, i.e. the next 90 top genes (in terms of variance to mean ratio of expression), for prior hyperparameters calibration for the optimal/suboptimal approaches. We follow the approach in \cite{dalton2011application} and employ the method of moments for prior calibration, but unlike \cite{dalton2011application}, a single set of hyperparameters is estimated and used for both clusters, since the labels of data are not available. It is well known that in small sample size settings, estimation of covariance matrices, scatter matrices and even mean
vectors may be problematic. Therefore, similar to \cite%
{dalton2011application}, we assume the following structure 
\begin{equation*}
\begin{split}
& \Psi _{0}=\Psi _{1}=%
\begin{bmatrix}
\sigma ^{2} & \rho \sigma ^{2} & \dots & \rho \sigma ^{2} \\ 
\rho \sigma ^{2} & \sigma ^{2} & \dots & \rho \sigma ^{2} \\ 
\vdots & \vdots & \ddots & \vdots \\ 
\rho \sigma ^{2} & \dots & \dots & \sigma ^{2}%
\end{bmatrix}%
_{d\times d}, \\
& m_{0}=m_{1}=m[1,\cdots ,1]_{d}^{T}, \\
& \nu _{0}=\nu _{1}=\nu ,\kappa _{0}=\kappa _{1}=\kappa ,
\end{split}%
\end{equation*}%
and estimate five scalars ($m$, $\sigma ^{2}$, $\rho $, $\kappa $, $\nu $)
from the data.

\begin{figure}[tph]
\centering
\includegraphics[width=0.45\textwidth]{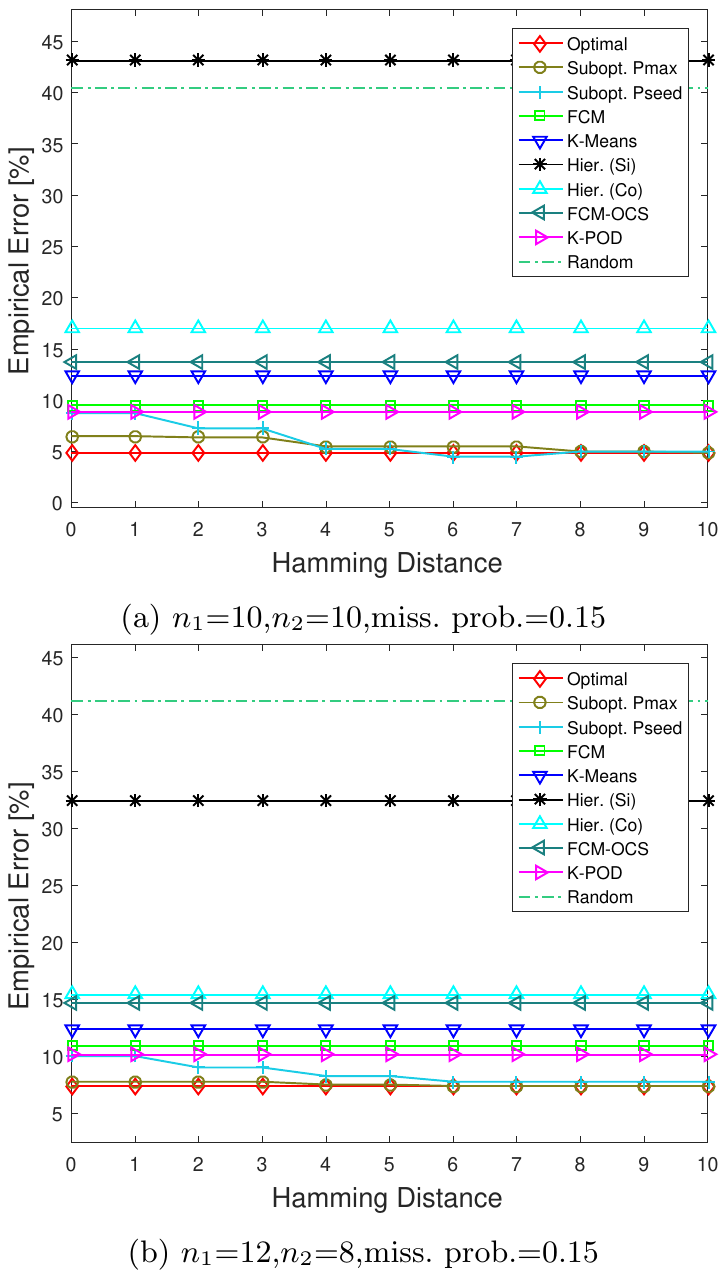}
\caption{\csentence{Empirical clustering errors on breast cancer RNA-seq data.}}
\label{fig:real}
\end{figure}

In each repetition, stratified sampling is done, i.e. $n_{1}$ and $n_{2}$
points are sampled randomly from each group (normal and tumor). When $%
n_{1}\neq n_{2}$, in half of the repetitions $n_{1}$ and $n_2$ points are randomly selected from the normal and tumor samples, respectively, and vice-versa in the other half. Prior calibration
is performed in each repetition, and 15\% of the selected features are considered as
missing values. Similar to the experiments on the simulated data, the clustering
error of each method in each iteration is calculated by comparing the
predicted labels and true labels of the sampled points (accounting for label
switching issue), and the average results over 40 repetitions are provided
in Figure \ref{fig:real}. It can be seen that the proposed optimal
clustering with missing values and its suboptimal versions outperform the
other algorithms. It is worth noting that the performance of Pseed is
more sensitive to the selected Hamming distance threshold for reference
partitions compared with the results on simulated data. 

\section*{Conclusion}

The methodology employed in this paper is very natural. As with any signal
processing problem, the basic problem is to find an optimal operator from a
class of operators given the underlying random process and a cost function,
which is often an error associated with operator performance. While it may
not be possible to compute the optimal operator, one can at least employ
suboptimal approximations to it while knowing the penalties associated with
the approximations.

In this paper, we have, in effect, confronted an old problem in signal
processing: If we wish to make a decision based on a noisy observed signal,
is it better to filter the observed signal and then determine the optimal
decision on the filtered signal, or to find the optimal decision based
directly on the observed signal? The answer is the latter. The reason is
that the latter approach is fully optimal relative to the actual observation
process, whereas, even if in the first approach the filtering is optimal
relative to the noise process, the first approach produces a composite of
two actions, filter and decision, each of which is only optimal relative to
a portion of the actual observation process. In the present situation
involving clustering, in the standard imputation-followed-by-clustering
approach, it is typically the case that neither the filter (imputation) nor
the decision (clustering) is optimal, so that even more advantage is
obtained by optimal clustering over the missing-value-adjusted RLPP.

\begin{backmatter}

\section*{List of abbreviations}
RLPP: random labeled point process;
RNA-seq: RNA sequencing;
MCAR: missing completely at random;
TCGA: The Cancer Genome Atlas;
FCM-OCS: fuzzy $c$-means with optimal completion strategy;
Hier. (Si): hierarchical clustering with single linkage;
Hier. (Co): hierarchical clustering with complete linkage;
GMM: Gaussian mixture model;
VST: variance stabilizing transformation;

\section*{Declarations}

\section*{Competing interests}
  The authors declare that they have no competing interests.

\section*{Author's contributions}
S. B. and S. Z. D. developed the method, performed the experiments, and wrote the first draft. X. Q. and E. R. D. proofread and edited the manuscript, and oversaw the project. All authors have read and approved final manuscript.

\section*{Availability of data and materials}
The publicly available real datasets analyzed during the current study have been generated by the TCGA Research Network https://cancergenome.nih.gov/.
\section*{Funding}
This work was funded in part by Awards CCF-1553281 and IIS-1812641 from the National Science Foundation, and a DMREF grant from the National Science Foundation, award number 1534534. The publication cost of this article was funded by Award IIS-1812641 from the National Science Foundation.
\section*{Ethics approval and consent to participate}
Not applicable.

\section*{Consent for publication}
Not applicable.

\section*{Acknowledgment}
We thank Texas A\&M High Performance Research Computing for providing
computational resources to perform experiments in this paper.

\bibliographystyle{bmc-mathphys} 
\bibliography{reference_final}

\section*{Additional file}
Additional file 1: Supplementary materials. Additional figures are given in a single multi-page PDF.

\end{backmatter}

\end{document}